\documentclass[journal]{IEEEtran}
\usepackage[utf8]{inputenc}
\usepackage{amssymb}
\usepackage{graphicx}
\usepackage{booktabs}
\usepackage{multirow}
\usepackage{cite}
\usepackage{subfigure}
\usepackage{enumerate}
\usepackage{threeparttable}
\usepackage[switch]{lineno} 
\usepackage{lipsum} 
\usepackage{colortbl}
\usepackage{hhline}
\usepackage{amsmath}
\usepackage{bm}

\DeclareMathOperator*{\argmin}{arg\,min}
\DeclareMathOperator*{\argmax}{arg\,max}
\newcommand{\cb}[1]{{\color{black} #1}}
\newcommand{\cc}[1]{{\color{blue} #1}}

\ifCLASSOPTIONcompsoc
  \usepackage[caption=false,font=normalsize,labelfont=sf,textfont=sf]{subfig}
\else
  \usepackage[caption=false,font=footnotesize]{subfig}
\fi
\begin{document}

\title{Active Ensemble Deep Learning for Polarimetric Synthetic Aperture Radar Image Classification}

\author{Sheng-Jie Liu$^\dagger$, Haowen Luo$^\dagger$, and Qian Shi
\thanks{$\dagger$ These authors contributed equally to this work.}
\thanks{Manuscript received Dec 10, 2019; accepted Jun 22, 2020. This work was supported by the National Natural Science Foundation of China under Grants 61601522 and 61976234. \emph{(Corresponding author: Qian Shi.)}} 
\thanks{S. J. Liu is with Department of Physics, University of Hong Kong, Hong Kong SAR, China (e-mail: sjliu.me@gmail.com).}
\thanks{H. Luo is with Department of
Geography and Resource Management, Chinese University of Hong Kong, Hong Kong SAR, China (e-mail:  hw.yue.luo@gmail.com).}
\thanks{Q. Shi is with Guangdong Provincial Key Laboratory of Urbanization and Geo-simulation, School of Geography and Planning, Sun Yat-sen University, Guangzhou 510275, China (e-mail: shixi5@mail.sysu.edu.cn).}
}

\markboth{Accepted by GRSL}%
{Accepted by GRSL}

\maketitle

\begin{abstract}
\cc{This is the preprint version. To read the final version
please go to \textit{IEEE Geoscience and Remote Sensing Letters} on
IEEE Xplore.}
Although deep learning has achieved great success in image classification tasks, its performance is subject to the quantity and quality of training samples. For classification of polarimetric
synthetic aperture radar (PolSAR) images, it is nearly impossible to annotate the images from visual interpretation. Therefore, it is urgent for remote sensing scientists to develop new techniques for PolSAR image classification under the condition of very few training samples. In this letter, we take the advantage of active learning and propose active ensemble deep learning (AEDL) for PolSAR image classification. We first show that only 35\% of the predicted labels of a deep learning model's snapshots near its convergence were exactly the same. The disagreement between snapshots is non-negligible.  From the perspective of multiview learning, the snapshots together serve as a good committee to evaluate the importance of unlabeled instances. Using the snapshots committee to give out the informativeness of unlabeled data, 
the proposed AEDL achieved better performance on two real PolSAR images compared with standard active learning strategies. It achieved the same classification accuracy with only 86\% and 55\% of the training samples compared with breaking ties active learning and random selection for the Flevoland dataset. 
\end{abstract}

\begin{IEEEkeywords}
Deep learning, convolutional neural network, active learning, ensemble learning, synthetic aperture radar, image classification
\end{IEEEkeywords}

\IEEEpeerreviewmaketitle

\section{Introduction}
Convolutional neural networks (CNNs) have obtained great successes in image classification tasks \cite{krizhevsky2012imagenet}, but the performance of CNNs is subject to the quantity and quality of training samples \cite{zhou2017fine}. 
When insufficient training samples are used to train a deep CNN, it tends to overfit on the training set and fails to achieve good machine generalization. 
Thus, existing state-of-the-art CNNs require large amounts of labeled data to achieve good performance. 
For instance, ImageNet, the most popular image dataset, consists of over 1 million images from daily life and is widely used to pretrain deep networks. 

For remote sensing image classification, it is difficult to obtain enough training data, especially when the satellite image is captured by radar remote sensing, for instance, synthetic aperture radar (SAR). Radar satellite images have a unique response to ground targets and are sensitive to the surface roughness and geometry \cite{beaudoin1990sar}, resulting in a very different image representation compared to optical images. 
As a result, it is nearly impossible to label radar images from visual interpretation in front of a computer. Moreover, collecting samples from field works requires expensive human labours  \cite{geng2017semisupervised}. Therefore, when using deep learning for PolSAR image classification, we are facing the dilemma between the need of sufficient training data and the limited resources for collecting reference data.

The image classification task with limited training data is well developed for hyperspectral image classification \cite{pan2018mugnet}. Several methods, including semisupervised learning \cite{ma2016semisupervised}, transfer learning \cite{camps2013advances} and active learning \cite{tuia2011survey}, have been proposed to tackle this problem. Among them, active learning aims at finding the most informative sample set to reduce the need of training data in supervised learning and has achieved promising results for hyperspectral image classification.  \cite{tuia2009active} first introduced maximum entropy (ME) criteria to select informative training samples in late 2000s. Later in early 2010s, mutual information \cite{mackay1992information} and breaking ties (BT) \cite{luo2005active} were also introduced to hyperspectral image classification  \cite{li2011hyperspectral}. Active learning strategies were further integrated with  transfer learning \cite{tuia2011using} and multiview learning \cite{zhou2016wavelet}. In the deep learning era, researchers extended their works to new deep learning methods with active learning on hyperspectral image classification \cite{liu2016active, haut2018active, liu2018wide}. 

\begin{figure*}[!t] 
\centering      
\includegraphics[width=0.75\textwidth]{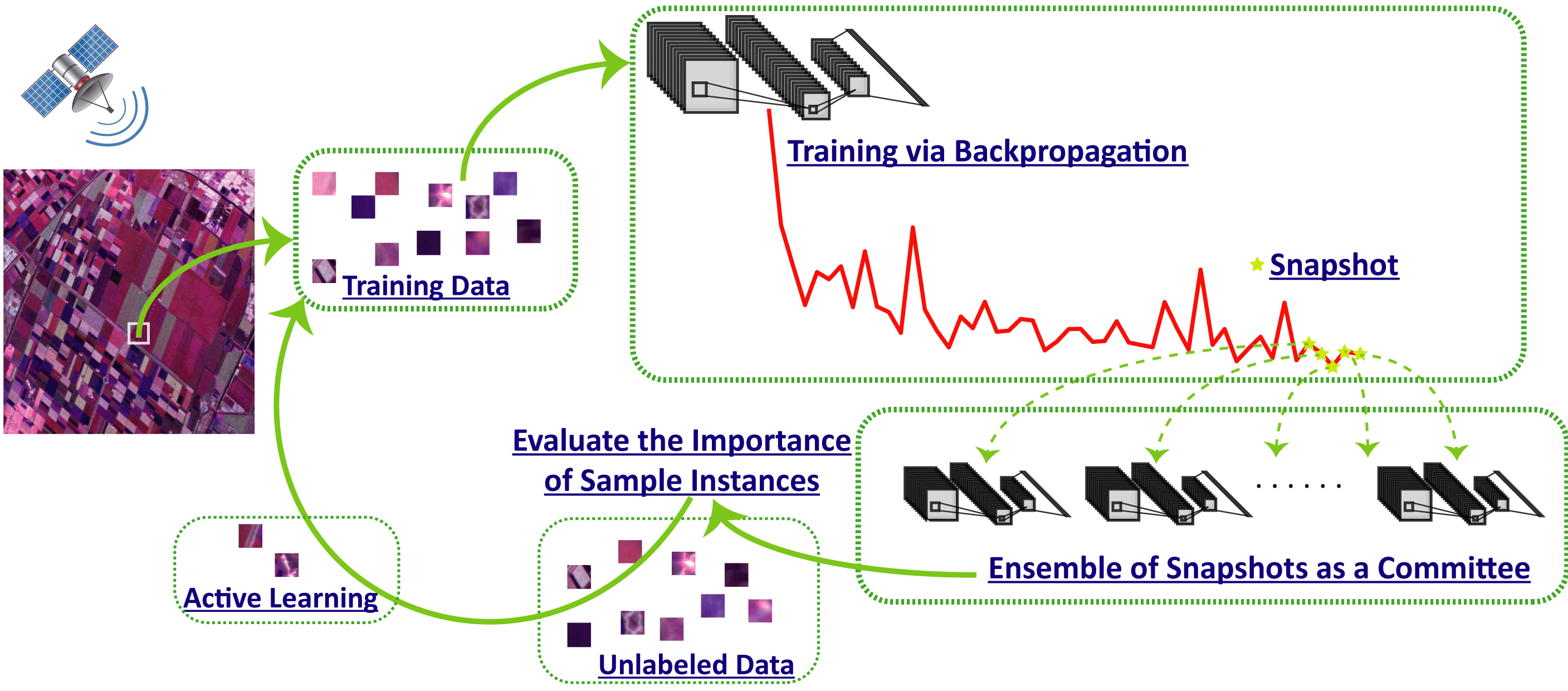}
\caption{Flowchart of the proposed method. We first use limited training data to train a deep learning model. Then, the model's snapshots near its convergence together serve as a committee to evaluate the importance of each samples instance. The most informative samples are added into the training set to further finetune the network until resources run out or the model's performance is satisfactory.}
\label{fig:flowchart}
\end{figure*}

Active learning with CNNs has shown promising results on hyperspectral image classification, but few attempted to use active learning on PolSAR images, which by the nature of data are in need of new classification methods under the context of very few training data. 
Exceptions are, a) \cite{liu2018novel} proposed an object-based method using random forest with active learning for PolSAR image classification; b) a recent study presented by \cite{bi2019active} integrates CNNs, active learning and Markov random field on PolSAR image classification. 
Given the lack of literature in this topic, in this study, our purpose is to develop new active deep learning methods for classification of PolSAR images.

Using backpropagation to find the local minima of CNNs can be traced back to late 1980s \cite{lecun1989backpropagation}, but it is time-consuming and difficult to find a set of optimal parameters for a CNN until the latest development of GPUs. Since AlexNet \cite{krizhevsky2012imagenet}, many optimizers are proposed to ease the pain of finding local minima for neural networks, e.g., Adam \cite{kingma2014adam}. To achieve robust and accurate results, \cite{huang2017snapshot} proposed a snapshot ensemble learning strategy to obtain ensemble of neural networks by converging  a single network to several local minima. For each convergence, they saved a set of parameters, which is referred to as a snapshot. The snapshots of a neural network can be regarded as a type of multiview learning.

As the key to active learning is to properly evaluate the informativeness of unlabeled instances, a comprehensive view from multiview learning can serve as a good committee and give out the importance of sample instances. In this letter, we propose a more straightforward active learning solution by using snapshots of a deep network near its convergence without resetting the learning rate to avoid additional training cost, as the disagreement near a network's convergence is still large enough to serve as a good committee.   The major contributions of this letter are summarized as follows.

a) We first show that even near the convergence of a deep learning model, the disagreement of its snapshots is still quite large. Using ensemble of neural networks can obtain a more robust and accurate result compared to single local minima.

b) As the disagreement of a network's snapshots is large enough, we propose a new active learning method under the context of deep neural networks (referred to as active ensemble deep learning, AEDL hereafter) for PolSAR image classification. The snapshots together serve as a committee to evaluate the informativeness of each sample instance. 

A flowchart is presented in Fig. \ref{fig:flowchart} illustrating the idea. After training a CNN with initial samples, the snapshots near its convergence serve as a committee to evaluate the importance of sample instances and add new actively selected training data in the next loop, until resources run out or the result is satisfactory.

\section{Methodology}
\label{sec:Methodology}
\subsection{Deep CNNs}
Deep CNNs have emerged as the dominant methods in the computer vision and pattern recognition community since 2012. The massive parameters of a deep CNN are the major characteristic that lead to its outstanding performance. Compared to fully-connected networks, deep CNNs contain at least one convolutional layer that can extract high-level spatial features. The output $y^{k}_{i,j}$  of the $k$th feature map for an image pixel located at position $(i,j)$ after convolving is given as
\begin{equation}
y^{k}_{i,j}=b^{k}+\sum_{m=1}^{M}\sum_{p=1}^{P}\sum_{q=1}^{Q}w^{k}_{p,q,m} \times x_{i+p,j+q,m},
\label{eq:conv} 
\end{equation}
where $P, Q$ are the height and width of the convolutional filters, $M$ is the number of input feature maps, and  $w^{k}_{p,q,m}$  and $b^{k}$ are the weight at $(p,q)$ of the convolutional filter and the bias  connected to the $m$th feature map. 

\subsection{Active Learning}
Active learning is a training strategy that iteratively enlarges the training set by querying samples from the unlabeled data.
The key to active learning is to select the most informative sample set with some criteria. 
In this study, three sampling schemes are considered under the context of AEDL, namely random selection, maximum entropy, and breaking ties.

\subsubsection{Random selection (RS)}
With RS strategy, the training set is enlarged by randomly selecting samples  from the unlabeled set. This strategy is independent to the classifier.

\subsubsection{Maximum entropy (ME)}
The ME active learning strategy was introduced to remote sensing image classification in \cite{tuia2009active}. This active learning method aims at finding the most informative sample set by maximizing the entropy of the probability values given an instance $x_i$ and an optimized classifier $\hat{\omega}$,
\begin{equation}
\hat{x}^{(ME)}_i = \argmax _{x_i,i\in U}H(y_i|x_i,\hat{\omega}),
\label{eq:entropy1} 
\end{equation}
and the entropy is formulated as,
\begin{equation}
H(y_i|x_i,\hat{\omega})=-\sum_k p(y_i=k|x_i,\hat{\omega})\log p(y_i=k|x_i,\hat{\omega}),
\label{eq:entropy2} 
\end{equation}
where $H(y_i|x_i,\hat{\omega})$ is the entropy of an instance $x_i$ given an optimized model  $\hat{\omega}$.

\subsubsection{Breaking ties (BT)}
The goal of BT active learning is to improve the probability difference between the best and second best classes. If these two probabilities are close, then BT aims at breaking the tie between them and thus improves classification confidence. The formula of BT is
\begin{equation}
\hat{x}^{(BT)}_i = \argmin_{x_i,i\in U}\{p\left(y_i=k_1|x_i,\hat{\omega}\right)-p\left(y_i=k_2|x_i,\hat{\omega}\right)\},
\label{eq:BT} 
\end{equation}
where $\hat{x}^{(BT)}_i$ denotes the selected informative samples,
$U$ is the unlabeled candidate set, $\hat{\omega}$ is the optimal parameters of the classifier, and $k_1$ and $k_2$ are the best and second best classes. 

\subsection{Snapshot Ensembles using Deep CNNs}
In this section, we recall the difference between conventional classifiers (e.g., support vector machines) and deep CNNs and show why ensemble of snapshots under the context of neural networks can achieve better generalization. The hinge loss of support vector machines as shown in  (\ref{eq:svm}) is a convex function and thus has unique global minima \cite{burges2000uniqueness}. 
For a support vector machine, suppose the hyperplane that separates two classes is $\mathbf{w}^T \mathbf{x}+b=0$,  where $\mathbf{w} \in \mathbb{R}^{k\times 1}$ is normal to the hyperplane and $\mathbf{x}$ is the sample instances, the optimization problem is 
\begin{equation}
\begin{aligned}
\mathop{\rm{min}}\limits_{\mathbf{w} \in \mathbb{R}^d}  \quad & \frac{1}{2}\|\mathbf{w}\|^2  \\
\textrm{s.t.}  \quad & y_i(\mathbf{w}^T\mathbf{x}_i+b) \geq  1 \quad \forall i. \\
\end{aligned}
\label{eq:svm}
\end{equation}
The optimal solution of the hyperplane $\mathbf{w}$ is unique.
However, neural networks are known as non-convex (except for a simple one-layer network) and the best way to solve this non-convex problem is gradient descent by the means of backpropagration.
There are multiple local minima in deep CNNs. Although one can find a local minima given a fixed training time, other snapshots of the model via backpropagation are also local minima given a specific condition and are acceptable solutions. We conducted an experiment (section \ref{ssec:disagreement}) showing that snapshots of a deep learning model are quite divergent when predicting labels of the unseen data. 
The ensemble of these snapshots can offer a more robust and accurate classification. 

\subsection{Active Ensemble Deep Learning}
Given that the snapshots of a deep learning model are  disagreed with one another, they naturally offer a comprehensive view to evaluate the unlabeled data under the umbrella of multiview learning. A committee from these snapshots should offer a more accurate assessment on the unlabeled data than a single ``optimal" snapshot. Therefore, we can base on snapshot ensembles and use active learning with deep CNNs to propose a new active learning strategy, the active ensemble deep learning (AEDL) method, as described earlier. 

Here we describe the BT active learning as an example under the context of AEDL. In the BT formula (\ref{eq:BT}), the probability of each instance $x_i$ is given under the condition of an optimal solution of the classifier parameters $\hat{\omega}$, which is unique if the classifier is a support vector machine or other convex classifiers. But under the context of deep learning, we know that the solution of $\hat{\omega}$ is only a local one. When we adopt the snapshot ensembles strategy, there exist several local minima and the parameters of snapshots ($\hat{\omega}_1, \hat{\omega}_2,..., \hat{\omega}_n$) are different with one another, where $n$ is the number of snapshots. In AEDL, the BT formula in (\ref{eq:BT}) is updated as, 
\begin{equation}
\begin{aligned}
\hat{x}^{(BT,AEDL)}_i = \argmin_{x_i,i\in U} \{p\left(y_i=k_1|x_i,\hat{\omega}_1,\hat{\omega}_2,..., \hat{\omega}_n\right) \\ 
- p\left(y_i=k_2|x_i,\hat{\omega}_1, \hat{\omega}_2,..., \hat{\omega}_n \right) \}. 
\end{aligned}
\label{eq:BT_AEDL} 
\end{equation}
Based on the theory of multiview learning, we know that $p\left(y_i=k|x_i,\hat{\omega}_1, \hat{\omega}_2,..., \hat{\omega}_n\right)$ is more robust and accurate than $p\left(y_i=k|x_i,\hat{\omega}\right)$. Therefore, the evaluated importance of unlabeled instances will be more accurate, resulting in a more robust and data efficient active learning environment.

\section{Experiments}
In this section, we first briefly introduce two real world PolSAR datasets used in the experiments. Then, we describe the experimental setup and show how divergent a network's snapshots could be. Finally, we present the results using the proposed AEDL on the two PolSAR datasets. 

\subsection{Datasets}

The first dataset used in the experiment is a L-band PolSAR dataset  captured by the NASA/JPL Airborne Synthetic Aperture Radar (AIRSAR) instrument. The image was captured over an agricultural coastal area of Flevoland, the Netherlands, in 1989, and is well recognized as a benchmark data for PolSAR image classification. The image contains 54,276 reference samples and 11 classes with a size of 512$\times$375. 
The Pauli composite and reference data are presented in \cb{the supplementary materials.}


The second dataset used in this study is a set of C,L,P-band PolSAR images captured by the AIRSAR instrument over another agricultural area in Flevoland, the Netherlands, in 1991. The image has a size of 1279$\times$1024 with 16 classes.
A L-band Pauli composite is presented \cb{the supplementary materials} with the reference data and corresponding legends.


\subsection{Experimental Setup}
We conducted the experiments on Python 3.6 using the TensorFlow framework with Keras as the high-level API. 
The wide contextual residual network \cite{liu2018wide} was adopted in the experiments and was accelerated by an Nvidia GTX1060 6G GPU. Before feeding samples into the network, we manually augmented the training samples four times by mirroring them across the horizontal, vertical and diagonal axes.
The Adam optimizer was used in backpropagation to train the network.
In active learning setting, we first trained the network from scratch for 100 epochs. Then, in each active learning, we combined the original training data and the actively selected training data to finetune the network for 30 epochs. The snapshots used in AEDL were then captured every two epochs. A total number of 9 snapshots were used in the experiments. In section \ref{ssec:analysis}, we analyze the effect of number of snapshots on the proposed AEDL. All reported results are averaged from 10 Monte Carlo runs.

\subsection{Disagreement of Deep Learning Models' Snapshots}
\label{ssec:disagreement}
\begin{figure}[htbp] 
\centering      
\includegraphics[width=0.4\textwidth]{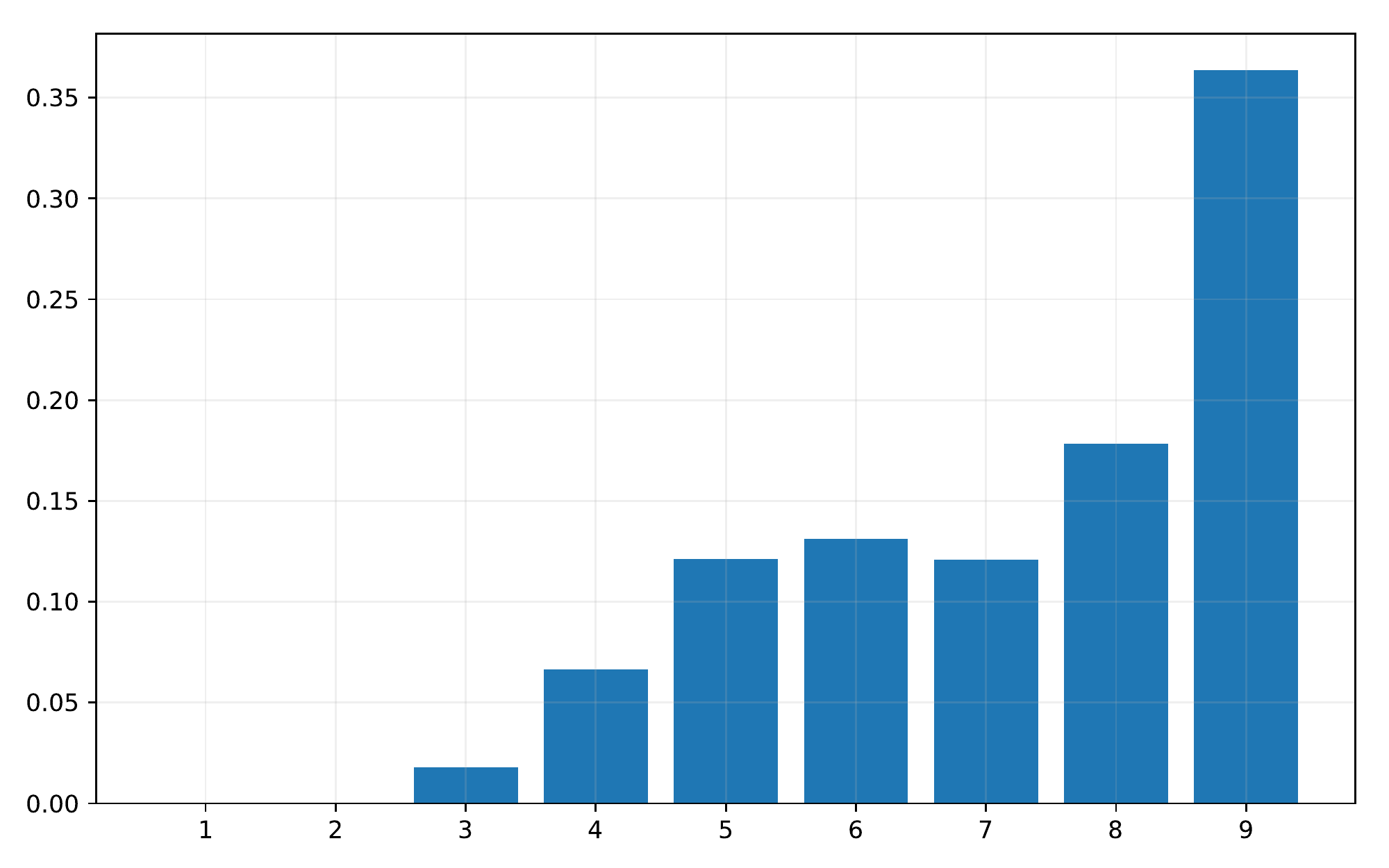}
\caption{Histogram of the number of the major predicted class.}
\label{fig:histogram}
\end{figure}

In the first experiment, we show how divergent one deep learning model's snapshots could be even near its convergence during training. We used the L-band Flevoland dataset as an example. Five samples per class were selected randomly to train the deep CNN. Nine snapshots were saved and used to predict the labels of the test set. Overall accuracies (OAs) of the snapshots were 70.32\%, 70.35\%, 64.59\%, 67.55\%, 70.64\%, 68.26\%, 69.90\%, 70.02\%, 70.65\%, and the ensemble OA was 71.44\%. We then show the histogram of number of the major predicted class in Fig. \ref{fig:histogram}.
Only 35\% of the predicted labels were exactly the same among the nine snapshots. About 20\% of the majority were not obtained overwhelmingly (less than 66\%). This illustrates the inconsistency of a deep CNN's snapshots even near its convergence, which on the other hand is the basis of multiview  learning.

\subsection{Active Learning on the L-band Flevoland Dataset}
\begin{figure}[htbp]  
\centering      
\includegraphics[width=0.4\textwidth]{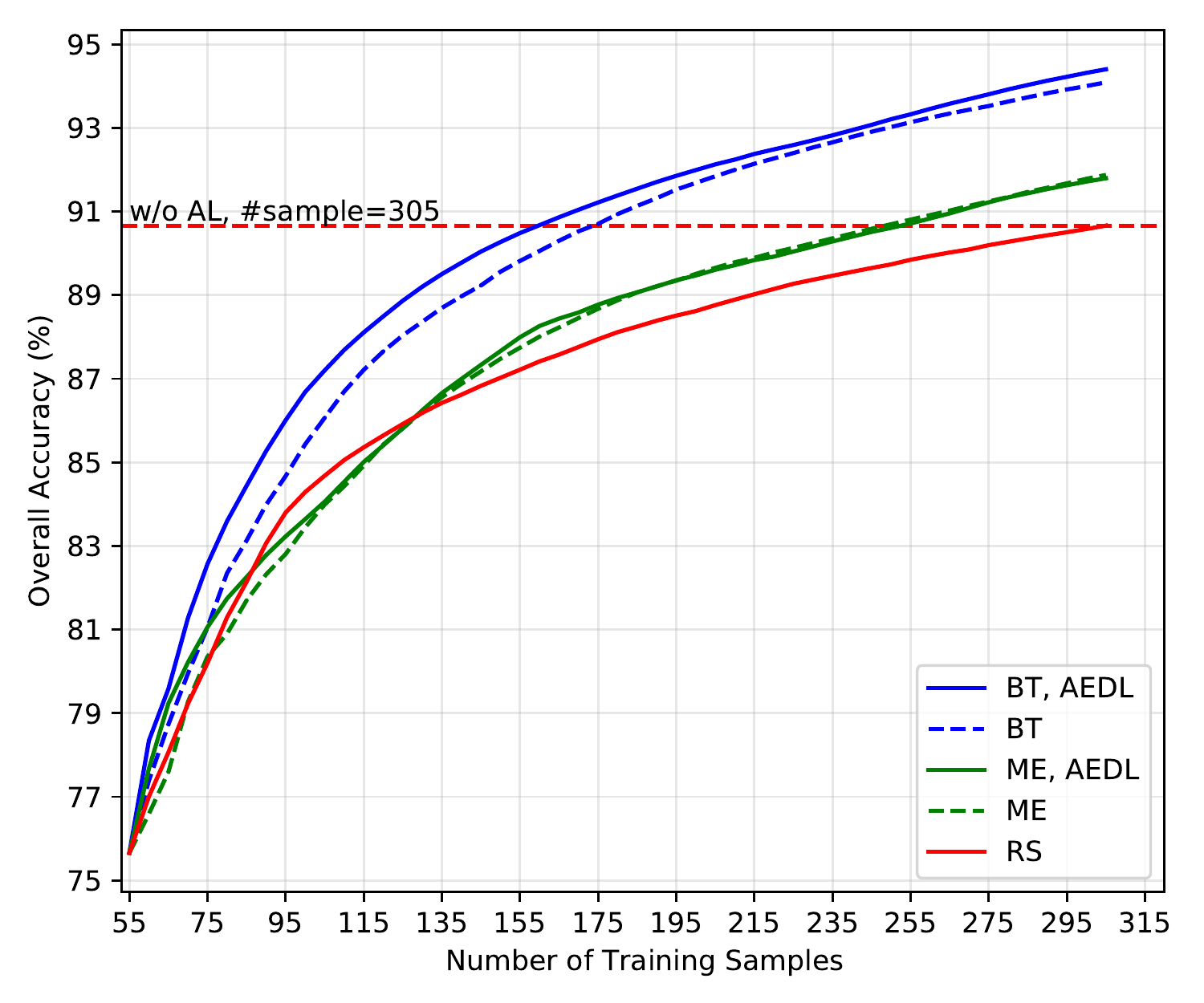}
\caption{Evaluation of overall accuracy against the number of training samples on the L-band Flevoland data.}
\label{fig:fle_AL}
\end{figure}

In the next experiment, we conducted active learning on the L-band Flevoland dataset. Beginning with 5 training samples per class, we randomly selected 5 samples per active learning from the candidates (20,000 samples), while the remaining 34,221 samples served as the test set. The result is presented in Fig.  \ref{fig:fle_AL}. When the size of training set was small, AEDL had a better performance than its competitors, leading a 1.1\% improvement using 115 samples for BT and a 0.4\% improvement using 75 samples for ME. 
Aiming at an OA of 89\%, AEDL used only 86\% and 55\% of the training data compared to BT and RS.

\subsection{Active Learning on the C,L,P-band Dataset}
\begin{figure}[htbp] 
\centering      
\includegraphics[width=0.4\textwidth]{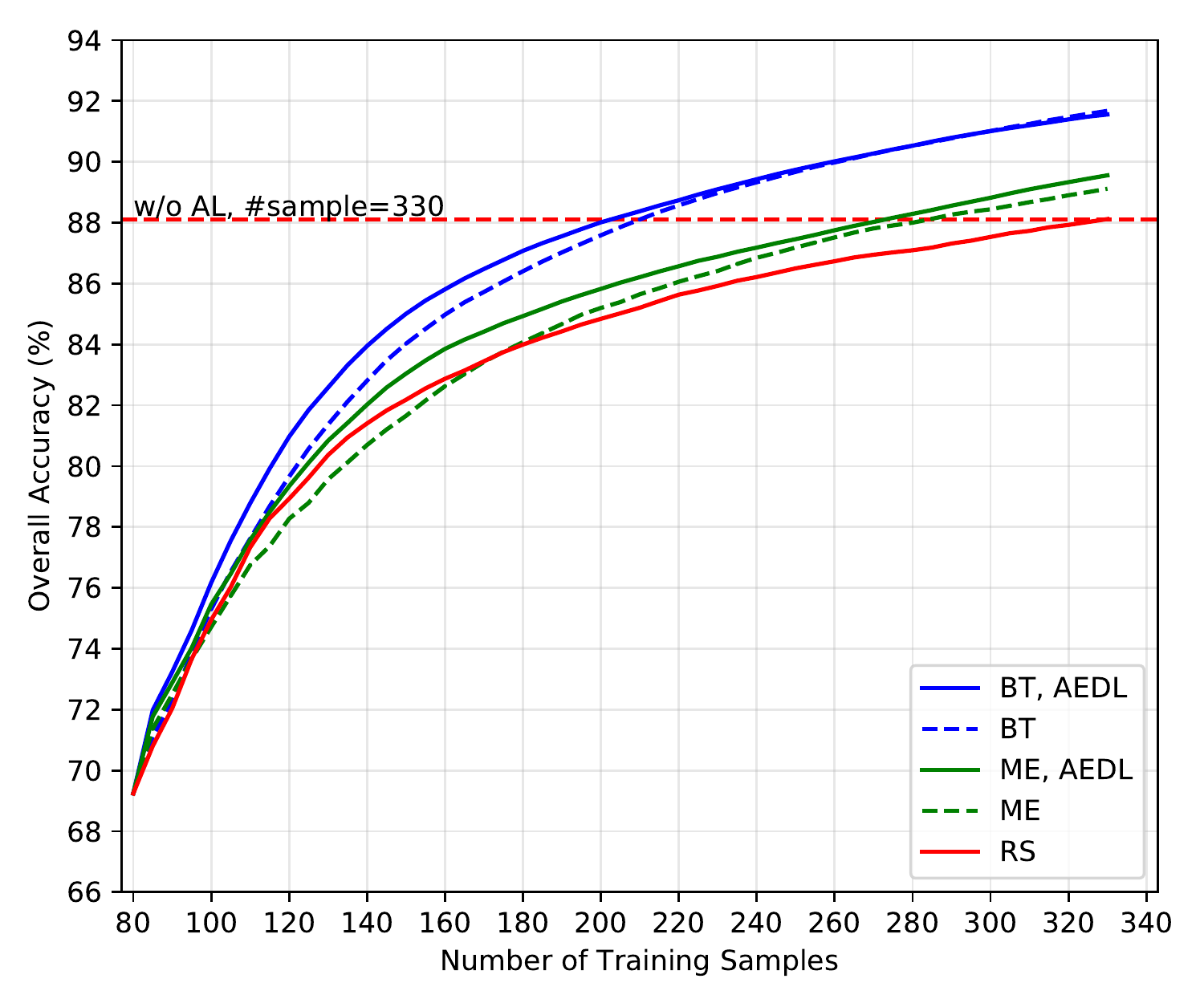}
\caption{Evaluation of overall accuracy against the number of training samples on the C,L,P-band Flevoland data.}
\label{fig:fle9clp_AL}
\end{figure}

Another PolSAR dataset, the C,L,P-band Flevoland captured in 1991, was used to test the generalization ability of AEDL. To facilitate the experiment, we randomly select 30,000 samples as candidates for active learning and 70,000 samples as the test set.
Beginning with 5 samples per class, we actively selected 5 samples in each active learning. The result is presented in Fig. \ref{fig:fle9clp_AL} and is similar to the one obtained from the L-band dataset. Both BT and ME with AEDL had a better performance than standard BT and ME. When the obtained samples were enlarged, the margin between AEDL and standard active learning became small, because the uncertainty of the snapshots became lower. The effectiveness of AEDL was maximized when the samples were extremely limited (120 samples), with an improvement of 0.7\% for BT and an improvement of 1.8\% for ME. The deep CNN obtained 88\% OA using the proposed AEDL-BT with only 200 samples, saving about 40\% of training samples compared to random selection (320 samples). 

\subsection{Analysis of the Sensitivity of the Number of Snapshots}
\label{ssec:analysis}
\begin{figure}[htbp] 
\centering      
\includegraphics[width=0.4\textwidth]{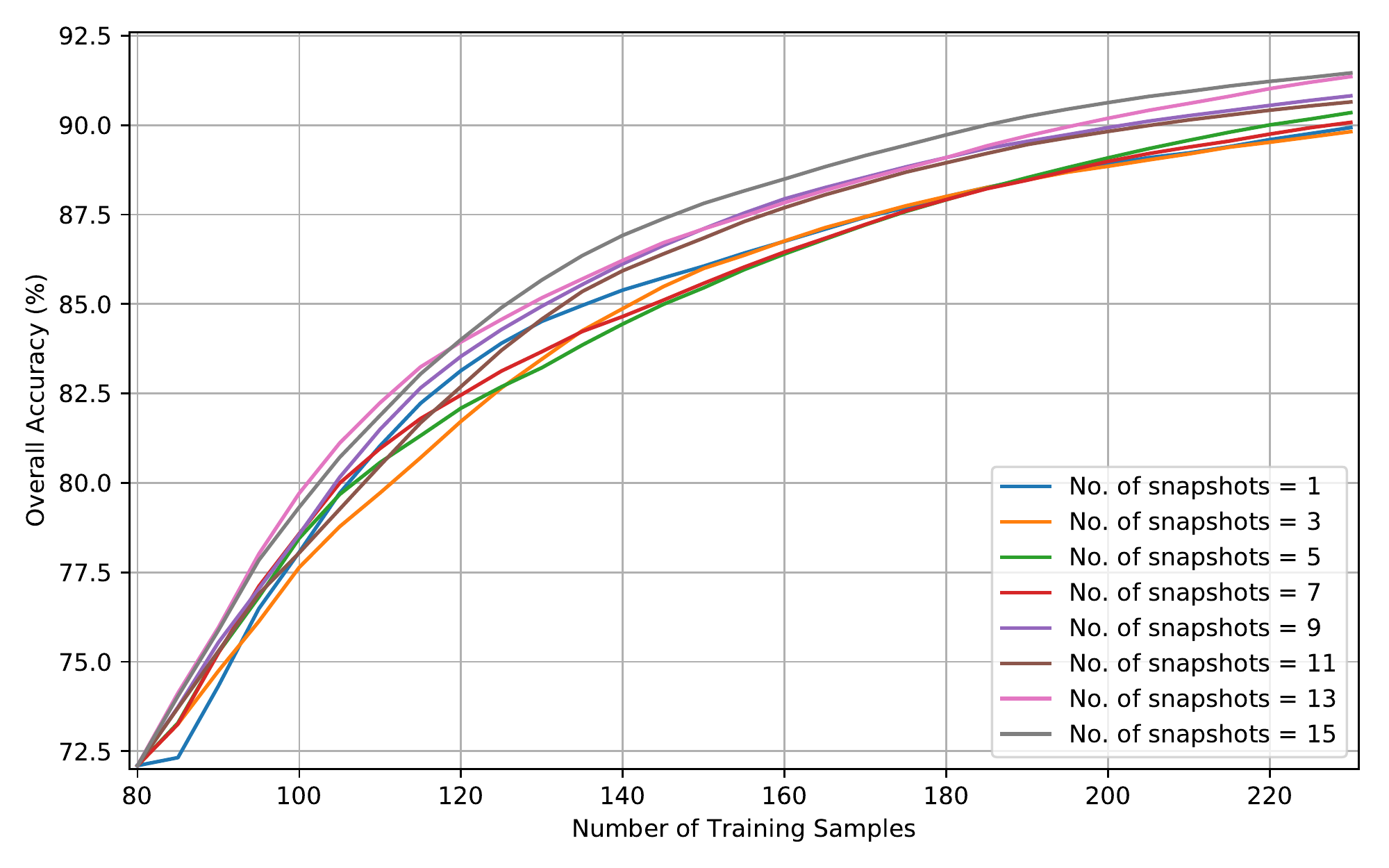}
\caption{Sensitivity analysis of the number of snapshots using the BT strategy.}
\label{fig:analysis}
\end{figure}

Finally, we conducted an experiment on the C,L,P-band Flevoland dataset to analyze the influence of the number of snapshots in AEDL with BT active learning and present the result in Fig. \ref{fig:analysis}. With the increase of snapshots in the committee, the performance of deep CNN with AEDL is better in terms of OA. The result is expected as a larger committee provides a more comprehensive view to evaluate the importance of sample instances.

\subsection{Generalization on other CNNs}
\begin{figure}[htbp] 
\centering  
\subfigure[DCCNN, L-band]{\includegraphics[width=0.24\textwidth]{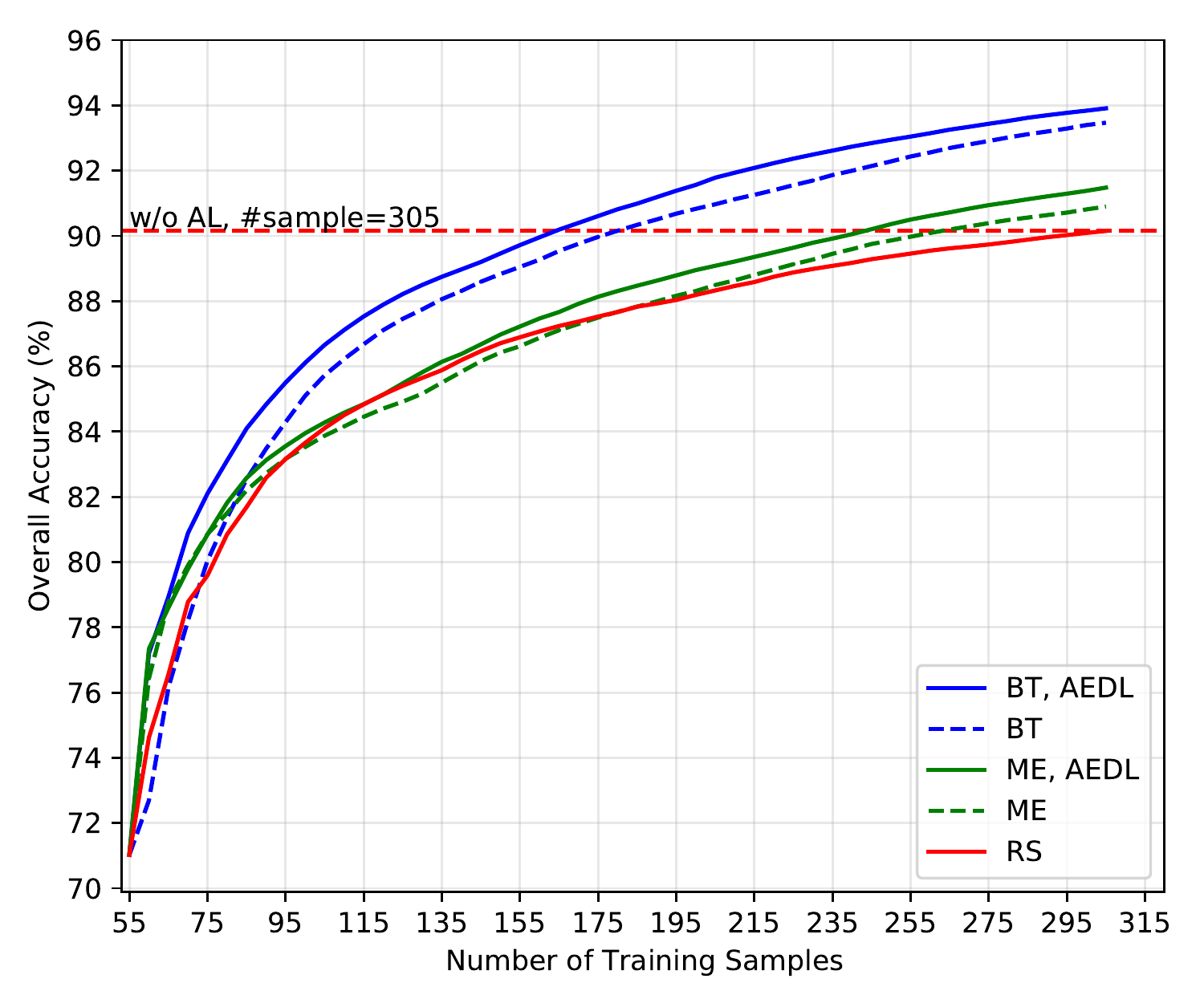}}
\subfigure[DCCNN, C,L,P-band]{\includegraphics[width=0.24\textwidth]{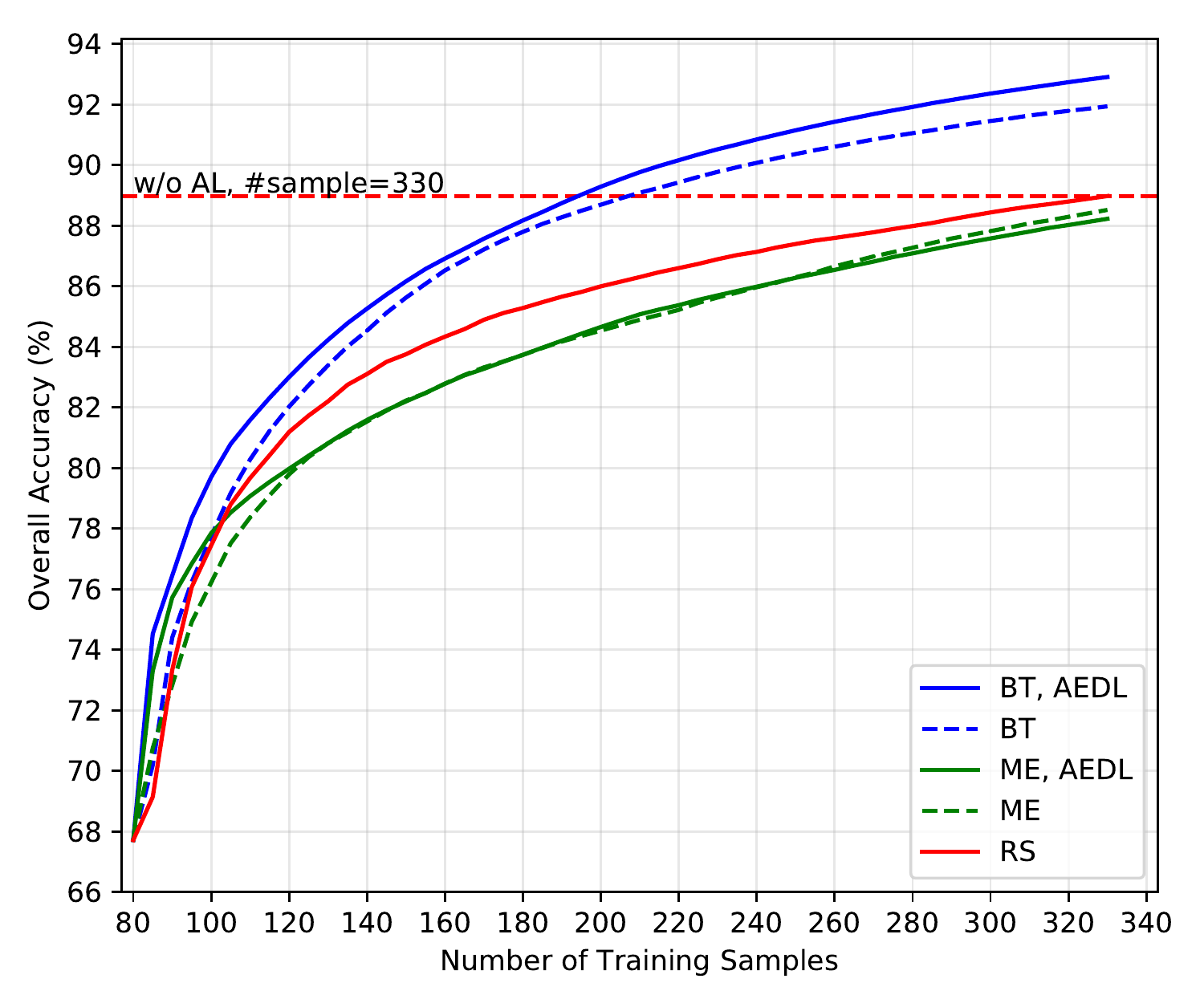}}
\subfigure[HResNet, L-band]{\includegraphics[width=0.24\textwidth]{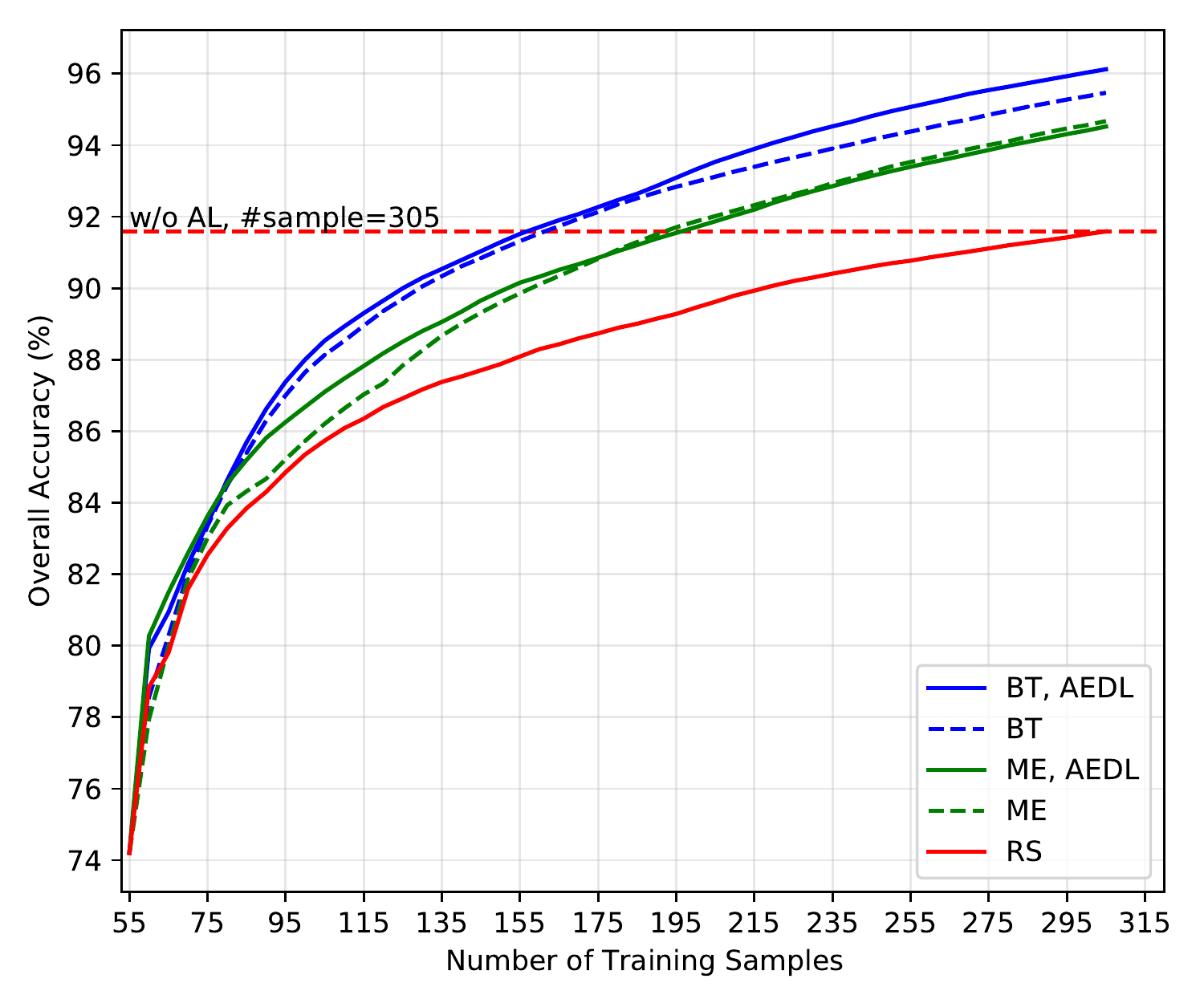}}
\subfigure[HResNet, C,L,P-band]{\includegraphics[width=0.24\textwidth]{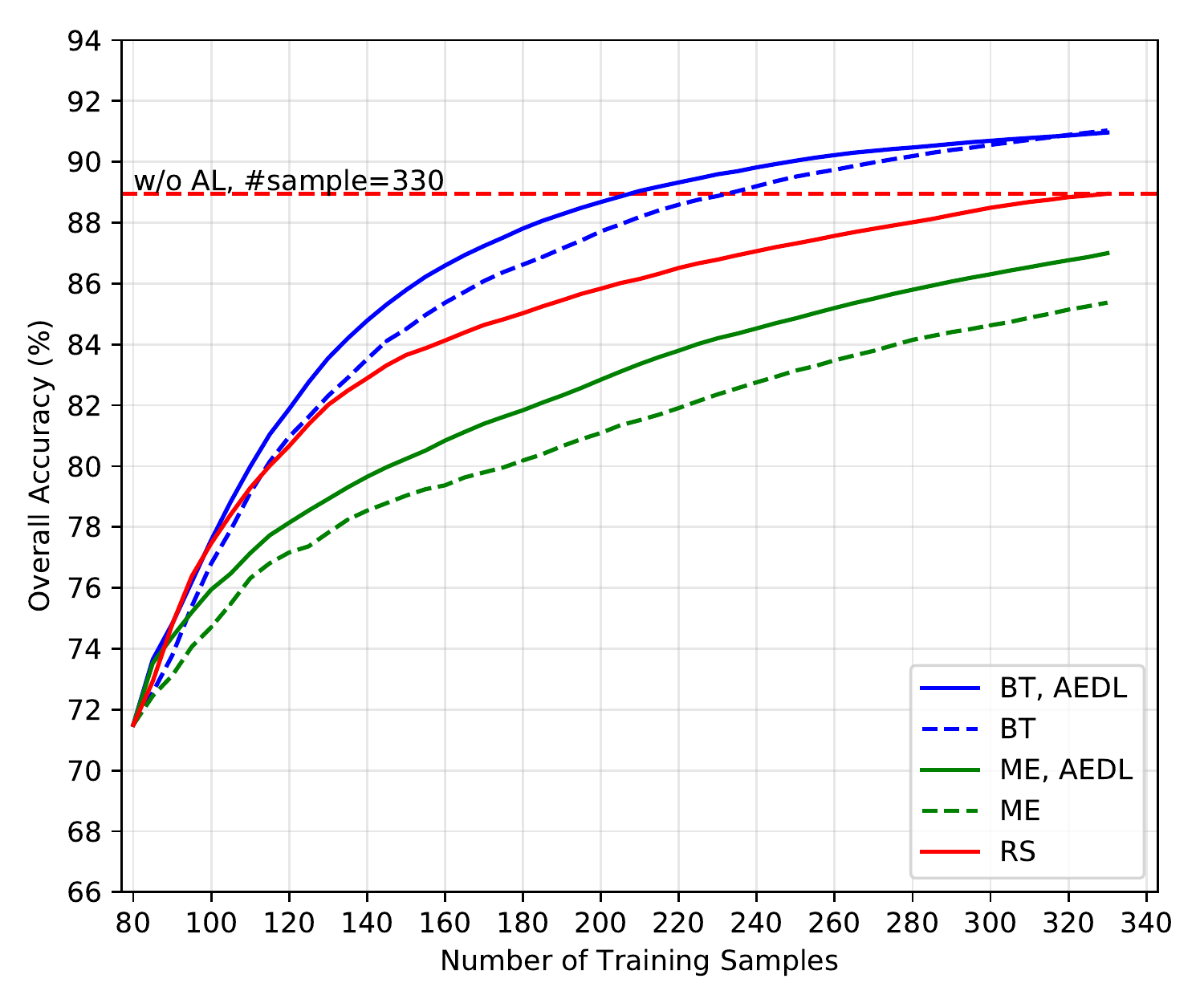}}
\caption{\cb{OA as a function of {\#}training samples using DCCNN and HResNet.}}
\label{fig:twocnns}
\end{figure}

\cb{We tested another two CNNs to show the generalization of the proposed method. The two additional CNNs are the deep contextual CNN (DCCNN) \cite{lee2017going} and the HResNet \cite{liu2020multitask}. Details of the networks can be found in the corresponding papers or supplementary materials. As shown in Fig. \ref{fig:twocnns},  the proposed method can enhance standard active learning methods using the two networks, indicating its generalization. 
    An interesting phenomenon is that with ME active learning, the networks performed worse than RS, which is in line with \cite{haut2018active}.
}

\section{Conclusion}
In this letter, we proposed active ensemble deep learning (AEDL) for  PolSAR image classification. 
The snapshots of a deep learning model near its convergence have a non-negligible disagreement and together provide a comprehensive view to evaluate the informativeness of sample instances. The obtained results demonstrate AEDL's promising ability to enhance PolSAR image classification with limited available data under the context of deep learning. 

\tiny{
\ifCLASSOPTIONcaptionsoff
  \newpage	
\fi
\bibliographystyle{IEEEtran}
\bibliography{strings}
}

\normalsize
\section*{Supplementary materials}
\subsection{Datasets}
\begin{figure}[htbp]
    \centering
    \subfigure[False-color Pauli RGB map]{\includegraphics[width=0.24\textwidth]{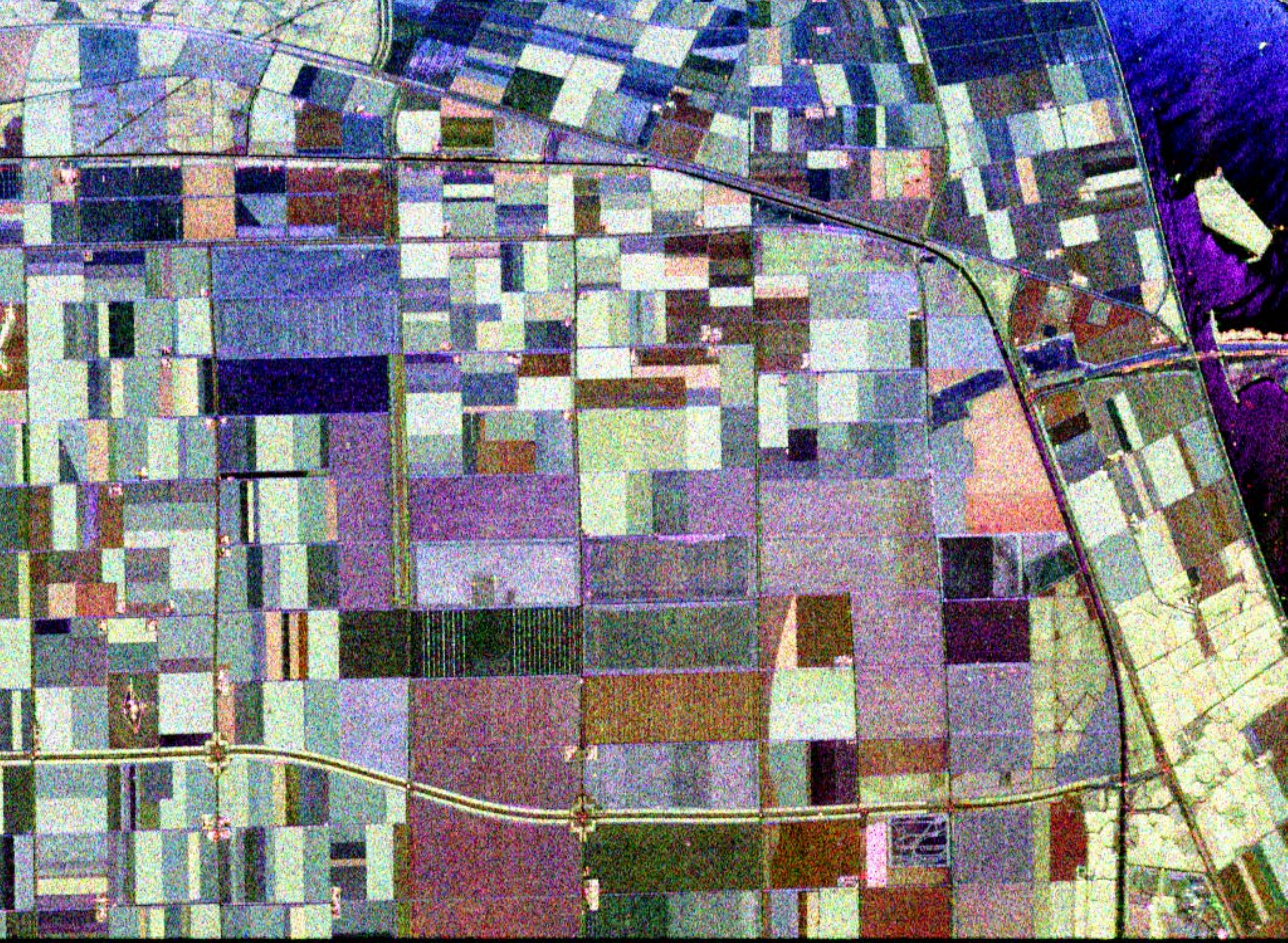}} 
    \subfigure[Ground truth]{\includegraphics[width=0.24\textwidth]{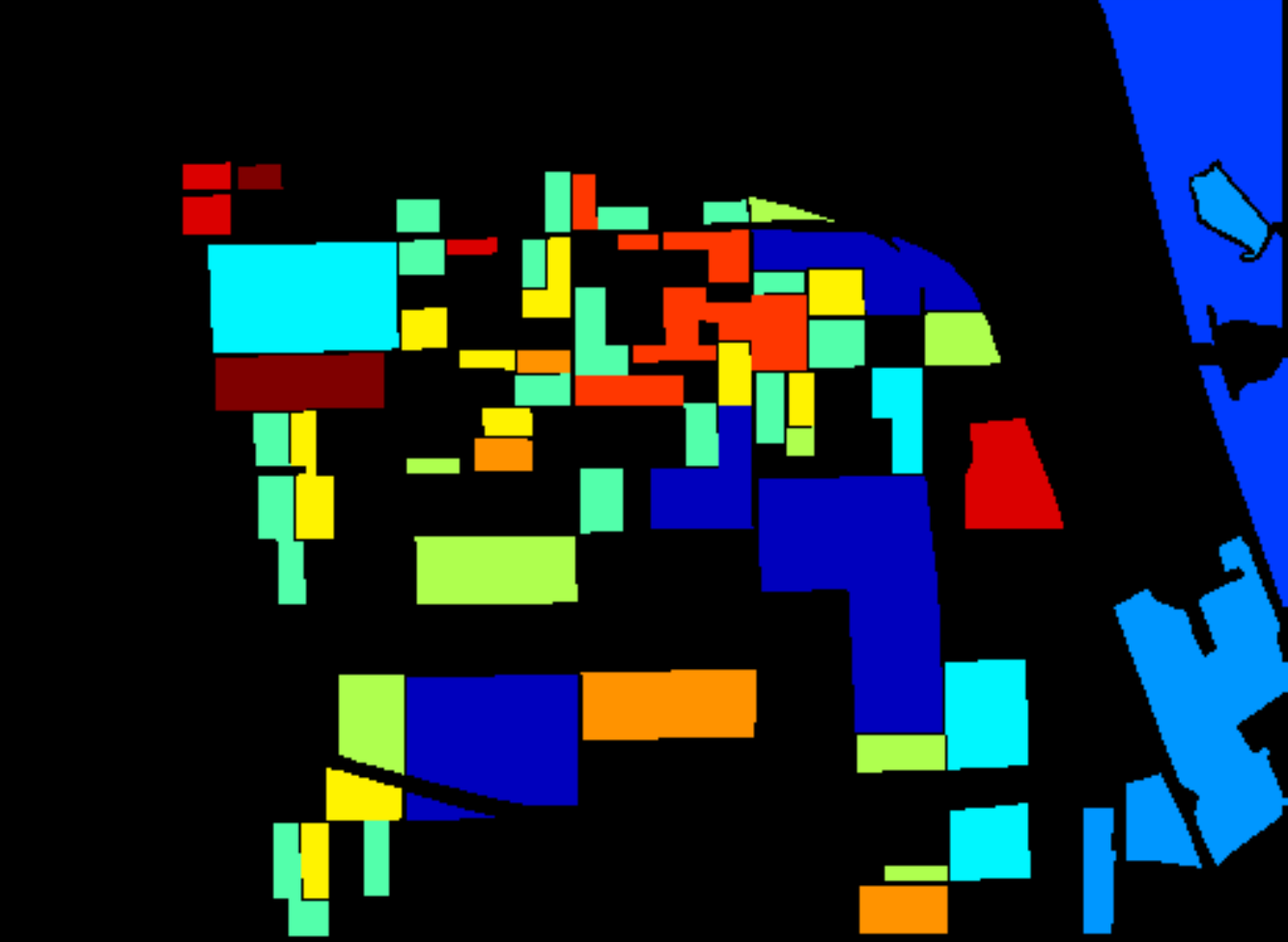}}
    \centering
    \subfigure{\includegraphics[width=0.48\textwidth]{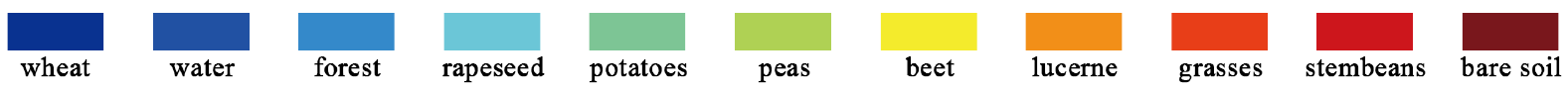}} 
    \caption{The L-band PolSAR image with reference data.}
    \label{fig:fle9}
\end{figure}

\begin{figure}[htbp]
    \centering
    \subfigure[L-band Pauli RGB map]{\includegraphics[width=0.21\textwidth]{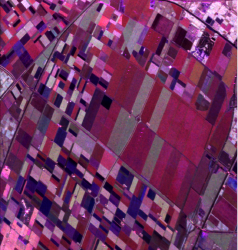}} 
    \subfigure[Ground truth]{\includegraphics[width=0.21\textwidth]{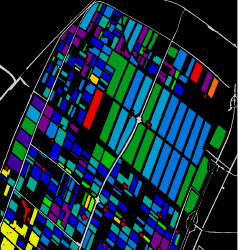}}
    \subfigure{\includegraphics[width=0.05\textwidth]{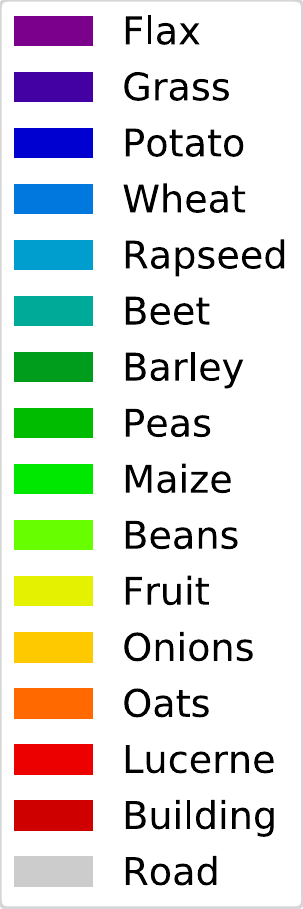}}
    \caption{The C,L,P-band PolSAR image with reference data.}
    \label{fig:fle9clp}
\end{figure}

Two real-world PolSAR datasets are used in the experiments, as shown in Figures \ref{fig:fle9} and \ref{fig:fle9clp}.

\subsection{Details of the three CNNs}
In the experiments, three CNNs are used to test the generalization performance of the proposed method. They are WCRN \cite{liu2018wide},  DCCNN \cite{lee2017going}, and HResNet \cite{liu2020multitask}. We show details of their architecture in Tables \ref{tab:wcrn}, \ref{tab:dccnn}, and \ref{tab:hresnet}, respectively. 

\newpage
\begin{table}[htbp]
  \centering
  \caption{WCRN's architecture.}
  \scalebox{0.9}[0.9]{
    \begin{tabular}{cccc}
    \toprule
    \toprule
    Index & Layer & Kernel Size & Output Size \\
    \midrule
    \midrule
    1a    & \multirow{2}[1]{*}{Multi-level Convolutions} & 1$\times$1$\times$64 & 5$\times$5$\times$64 \\
    1b    &       & 3$\times$3$\times$64 & 3$\times$3$\times$64 \\
    2a    & MaxPooling(1a) & 5$\times$5   & 1$\times$1$\times$64 \\
    2b    & MaxPooling(1b) & 3$\times$3   & 1$\times$1$\times$64 \\
    3     & Concatenate(2a,2b) & -     &  \\
    4     & BN+ReLU+Conv & 1$\times$1$\times$128 & 1$\times$1$\times$128 \\
    5     & BN+ReLU+Conv & 1$\times$1$\times$128 & 1$\times$1$\times$128 \\
    6     & Add(3,5) & -     & 1$\times$1$\times$128 \\
    7     & Flatten + FC + SoftMax & \#Class & \#Class \\
    \bottomrule
    \bottomrule
    \end{tabular}}%
  \label{tab:wcrn}%
\end{table}%

\begin{table}[htbp]
  \centering
  \caption{DCCNN's architecture.}
  \scalebox{0.9}[0.9]{
    \begin{tabular}{cccc}
    \toprule
    \toprule
    Index & Layer & Kernel Size & Output Size \\
    \midrule
    \midrule
    1a    & \multirow{3}[1]{*}{Multi-level convolutions} & 1$\times$1$\times$128 & 5$\times$5$\times$128 \\
    1b    &       & 3$\times$3$\times$128 & 3$\times$3$\times$128 \\
    1c    &       & 5$\times$5$\times$128 & 1$\times$1$\times$128 \\
    2a    & MaxPooling(1a) & 5$\times$5   & 1$\times$1$\times$384 \\
    2b    & MaxPooling(1b) & 3$\times$3   & 1$\times$1$\times$384 \\
    3     & Concatenate(2a,2b,1c) & -     & 1$\times$1$\times$384 \\
    4     & ReLU+BN+Conv & 1$\times$1$\times$128 & 1$\times$1$\times$128 \\
    5     & ReLU+BN+Conv & 1$\times$1$\times$128 & 1$\times$1$\times$128 \\
    6     & ReLU+Conv & 1$\times$1$\times$128 & 1$\times$1$\times$128 \\
    7     & Add(4,6) & -     & 1$\times$1$\times$128 \\
    8     & ReLU+Conv & 1$\times$1$\times$128 & 1$\times$1$\times$128 \\
    9     & ReLU+Conv & 1$\times$1$\times$128 & 1$\times$1$\times$128 \\
    10    & Add(7,9) & -     & 1$\times$1$\times$128 \\
    11    & ReLU+Conv+ReLU+Dropout(0.5) & 1$\times$1$\times$128 & 1$\times$1$\times$128 \\
    12    & Conv+ReLU+Dropout(0.5)+Conv & 1$\times$1$\times$128 & 1$\times$1$\times$128 \\
    13    & Flatten + FC + SoftMax & \#Class & \#Class \\
    \bottomrule
    \bottomrule
    \end{tabular}}%
  \label{tab:dccnn}%
\end{table}%

\begin{table}[htbp]
  \centering
  \caption{HResNet's architecture.}
  \scalebox{0.9}[0.9]{
    \begin{tabular}{cccc}
    \toprule
    \toprule
    Index & Layer & Kernel Size & Output Size \\
    \midrule
    \midrule
    1     & Conv  & 3$\times$3$\times$64 & 7$\times$7$\times$64 \\
    2     & BN+ReLU+Conv & 3$\times$3$\times$64 & 7$\times$7$\times$64 \\
    3     & ReLU+Conv & 3$\times$3$\times$64 & 7$\times$7$\times$64 \\
    4     & Add(1,3) & -     & 7$\times$7$\times$64 \\
    5     & Global Average Pooling & 7$\times$7   & 64 \\
    6     & FC + SoftMax & \#Class & \#Class \\
    \bottomrule
    \bottomrule
    \end{tabular}}%
  \label{tab:hresnet}%
\end{table}%

\end{document}